\pdfoutput=1
\documentclass[11pt]{article}

\usepackage[]{acl}
\usepackage{url}
\usepackage{times}
\usepackage{latexsym}
\usepackage{graphicx}
\usepackage{subcaption}

\usepackage{algorithm}
\usepackage[noend]{algpseudocode}
\makeatletter
\def\BState{\State\hskip-\ALG@thistlm}
\makeatother
\usepackage[T1]{fontenc}
\usepackage[utf8]{inputenc}

\usepackage{microtype}

\usepackage{amsthm}
\theoremstyle{definition}
\newtheorem{example}{Example}

\title{The Effectiveness of Masked Language Modeling and Adapters for Factual Knowledge Injection}

\author{Sondre Wold \\ University of Oslo}

\begin{document}
\maketitle
\begin{abstract}
This paper studies the problem of injecting factual knowledge into large pre-trained language models. We train adapter modules on parts of the ConceptNet knowledge graph using the masked language modeling objective and evaluate the success of the method by a series of probing experiments on the \textsc{lama} probe. Mean P@K curves for different configurations indicate that the technique is effective, increasing the performance on subsets of the \textsc{lama} probe for large values of \textit{k} by adding as little as 2.1\% additional parameters to the original models.
\end{abstract}

\section{Introduction}
Large pre-trained language models (PLMs) are difficult to interpret due to their complexity and large parameter size. This can partly be explained by the nature of popular training regimens, such as the masked language modelling objective, which encodes distributional knowledge. Such regimens have proven effective for a range of downstream NLP tasks, but they also make it difficult to determine and validate the origin of whatever knowledge the models end up with.

Consequently, there have been multiple efforts to integrate structured information into PLMs \citep{peters-etal-2019-knowledge, yasunaga-etal-2021-qa, kaur-etal-2022-lm}. This has not only been motivated by the promise of better interpretability, but also the observation that there exist scenarios where we would want to stress information that might not be so easily encoded by modelling long range dependencies between fragments of text. This includes knowledge intensive tasks where employing the correct factual knowledge is crucial, for example within the medical domain \cite{zhang-etal-2021-smedbert} and question answering \citep{zhang2022greaselm}. At the same time, there exist multiple structured sources that attempt to capture factual knowledge. These sources range from domain specific knowledge graphs for medical information \citep{shi2017semantic}, commonsense graphs like Yago or ConceptNet \citep{suchanek2007yago,speer2017conceptnet}, to lexico-semantic networks like WordNet \citep{miller1995wordnet}.

In this paper, we attempt to inject the structured information found in the ConceptNet knowledge graph \citep{speer2017conceptnet} into pre-trained language models. The injection is done by training relatively small neural networks, known as adapter modules \citep{houlsby2019parameterefficient, pfeiffer-etal-2020-adapterhub}, on subject---predicate---object triples. As in \citet{lauscher2020common}, we extract the triples using a random walk procedure and then translate them into natural language so that we can use masked language modeling as the training objective. The resulting adapters are injected into all layers of two popular pre-trained language models: \textsc{bert} base \citep{devlin-etal-2019-bert} and \textsc{RoBERTa} base \citep{liu2019roberta}. Our code and data is made publicly available\footnote{\url{https://github.com/SondreWold/adapters-mlm-injection}}.

For the injection to be deemed effective, we argue that the adapter-injected models must be able to use the knowledge gained from the adapter training together with what the models learned during their initial pre-training. In order to quantitatively assess this, we evaluate our models in a zero-shot setting on the ConceptNet subset of the \textsc{lama} probe \citep{petroni2019language}. As ConceptNet is the source for both our training corpus and the \textsc{lama} probe, we can better measure how much of the factual knowledge seen during adapter training the models can be expected to recall. 

\section{Related work}
\begin{figure}
	\begin{center}
		\includegraphics[width=0.5\textwidth]{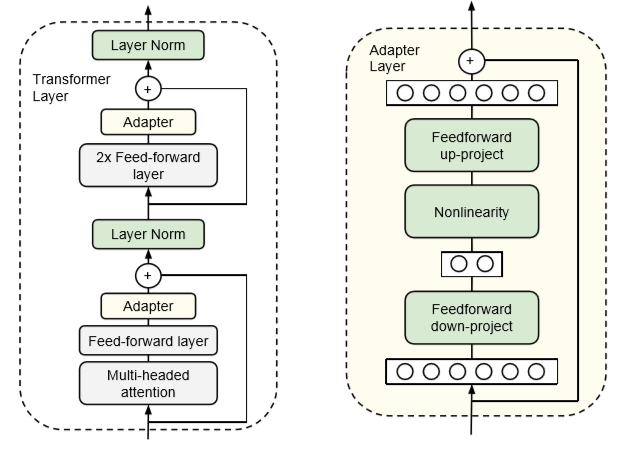}
	\end{center}
	\caption{Left: how adapters are injected into each transformer layer. Right: the components of each adapter module. Figure from \citet{houlsby2019parameterefficient}.}
	\label{fig:houlsbyAdapter}
\end{figure}
Combining structured information with language models is a standing problem in NLP. One approach to overcome this has been to combine knowledge graphs with PLMs, augmenting the distributional knowledge encoded in the models with the structured information found in the graphs \cite{sun2021jointlk, Liu_Zhou_Zhao_Wang_Ju_Deng_Wang_2020, wang2020kadapter}. Within this approach, we find several uses of adapters. First introduced for NLP by \citet{houlsby2019parameterefficient}, and popularized by the AdapterHub framework \citet{pfeiffer-etal-2020-adapterhub}, adapters are small neural networks injected into larger, often pre-trained models. During training the original model weights are kept static, and only the set of newly introduced weights from the adapter are adjusted. Figure \ref{fig:houlsbyAdapter} illustrates the architecture proposed by \citet{houlsby2019parameterefficient} and how it is injected into a transformer layer. 

The methodology in this paper is inspired by \citet{lauscher2020common}, who inject commonsense information and world knowledge into \textsc{bert} by using such adapter modules. As in our work, the adapters train with the masked language modeling objective over subject---predicate---object triples from the ConceptNet graph, but they are evaluated on the \textsc{glue} benchmark \citep{wang2018glue}. Although the result are inconclusive for most of the tasks in \textsc{glue}, the injected models perform better than their base model counterparts on the world knowledge and commonsense categories of the diagnostic set.

A similar approach is taken by \citet{wang2020kadapter}. Their \textsc{k-dapter} model has one adapter for factual knowledge, trained on aligned text triplets from Wikipedia and Wikidata, and one for linguistic knowledge, obtained via dependency parsing. Results on knowledge-driven
tasks, including relation classification, entity
typing, and question answering, show that this setup improves performance, and furthermore, that \textsc{K-adapter} captures more versatile knowledge than \textsc{RoBERTa}.

In a more domain specific context, \citet{meng-etal-2021-mixture} use adapter modules to infuse a large biomedical knowledge graph into an underlying \textsc{bert} model. By partitioning the large graph into smaller sub-graphs, which are then fed into distinct adapter modules and fused using a mixture layer that combine the knowledge from all the adapters using an attention layer, they achieve a new state-of-the-art performance on five domain specific datasets.

\section{Experiments}

Following \citet{lauscher2020common}, we use the same configuration for our adapter modules as in \citet{houlsby2019parameterefficient}. We set the size of the adapter modules to 64, which implies a reduction factor of 12 from the original transformer layer size of 768 in \textsc{bert}\textsubscript{\textsc{base}}. This increases the total amount of parameters by 2.1\%.  We use \textsc{GELU} \citep{hendrycks2020gaussian} as the activation function inside the adapters, and the Adam optimizer from \citep{kingma2017adam}. We set the learning rate to \textit{1e-4} with $10.000$ warm-up steps and weight decay factor of  $0.01$.  We allow the adapter to train for $100.000$ optimization steps while freezing all the original transformer weights. The adapters are implemented using the \texttt{adapter-transformers} library \citep{pfeiffer-etal-2020-adapterhub}. Throughout the remainder of this paper, the resulting configuration is referred to as \textsc{CN}\textsubscript{\textsc{Houlsby 100k}} in figures and as the Houlsby configuration in text. 

The adapters train on the same subset of ConceptNet as in \citet{lauscher2020common}. As this study was named Retrograph, we refer to this particular set of predicate types as the Retrograph predicate set. The predicates in this set are: \textsc{antonymOf}, \textsc{synonymOf}, \textsc{IsA} and \textsc{mannerOf}. Subject---predicate---object triplets with one of these predicates in their middle position are extracted through a random traversal procedure\footnote{Details on this traversal procedure can be found in \citet{lauscher2020common} or in appendix \ref{sec:appendix}.} and then subsequently chained so that we get blocks of text in natural language on the following format: \\
\begin{example}
	\item \label{ex:corpusExample} 
	possible is a synonym of possibility.  \\
	possibility is a concept. \\
	concept is a synonym of conception.\\
	conception is a synonym of fertilization.\\
	fertilization is a enrichment.\\
	enrichment is a gift.\\
\end{example}

The corpus is processed using masked language modeling (MLM), parsed line for line with a MLM probability of $0.15$, as in the original \textsc{bert} paper \citep{devlin-etal-2019-bert}. We also experiment by training on the corpus by a maximum sequence length instead of line by line training. However, this did not affect the performance of the models in any significant way. 

\subsection{Evaluation}

We evaluate our injected models on the ConceptNet split of the \textsc{lama} (LAnguage Model Analysis) probe \citep{petroni2019language}, which allows for testing of the factual and commonsense knowledge of language models. Facts are presented as fill-in-the-blank cloze statements, e.g: "Ibsen was born in [\textsc{mask}] in the year 1828", and models are ranked based on how highly it ranks the ground truth token. All models are evaluated in a zero-shot setting, using the same prediction head as in their pre-training.  

As we train our adapter modules on ConceptNet and also evaluate on the ConceptNet split from \textsc{lama}, it is important to note that what we test here is not the model's ability to generalize on unseen data in the traditional sense, but whether or not they are able to reproduce the factual information extracted from the knowledge graph during adapter training. The phrasing of the cloze statements in \textsc{lama} is not the same as in the training corpus for the adapters, although fairly similar. For example, one sentence in LAMA derived from the source triple \texttt{communicating hasSubevent knowledge} is presented in the probe as \textit{Communicating is for gaining [MASK]}, while the same triple would be phrased as \textit{communication has subevent knowledge} in the training corpus for the adapters. This makes it possible to control the degree of overlap between instances of factual knowledge in the training corpus and the concepts at the object position in the statements from \textsc{lama}. The degree of overlap is numerically specified in the discussion of each result. 

\subsubsection{Evaluation metric}
Following \citet{petroni2019language}, we use mean precision at different values of \textit{k} as the evaluation metric over the \textsc{lama} resource. Normally, as in information retrieval, we calculate the precision of a retrieved collection as the number of relevant documents proportionate to the total number of retrieved documents. Here, however, we only have one true positive for collections of all sizes. Thus, the mean precision at various values for $k$ is equal to the whether or not the correct word is a member of the set of predictions of size $k$. If $k=100$, we return a precision of $1$ if the correct word is one of the top $100$ predictions. 

\section{Results}
\begin{figure}
	\centering
	\subfloat[\centering \textsc{All predicates} ]{{\includegraphics[width=5.5cm]{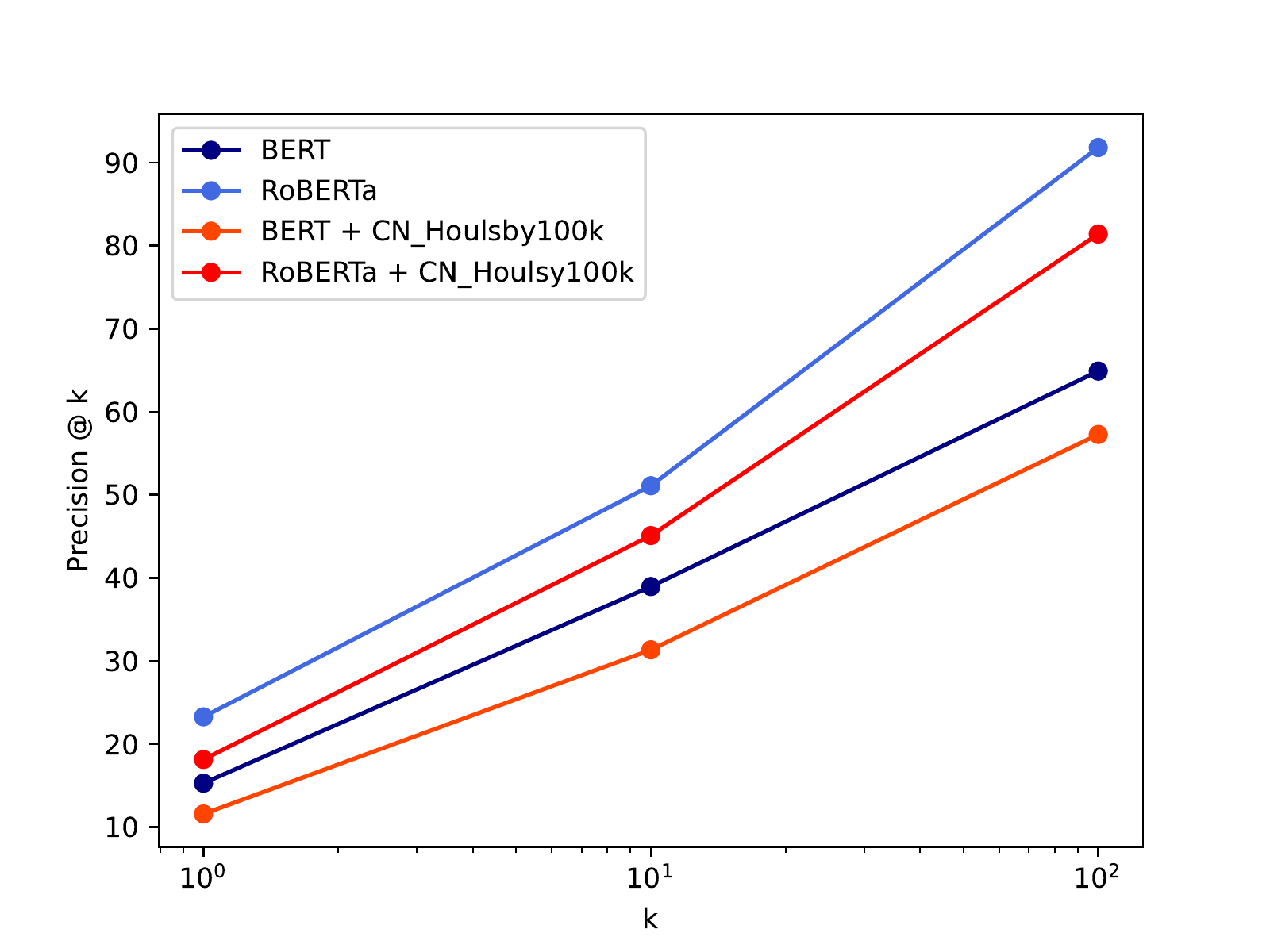}}}%
	\qquad
	\subfloat[\centering \textsc{the \textsc{isA} predicate} ]{{\includegraphics[width=5.5cm]{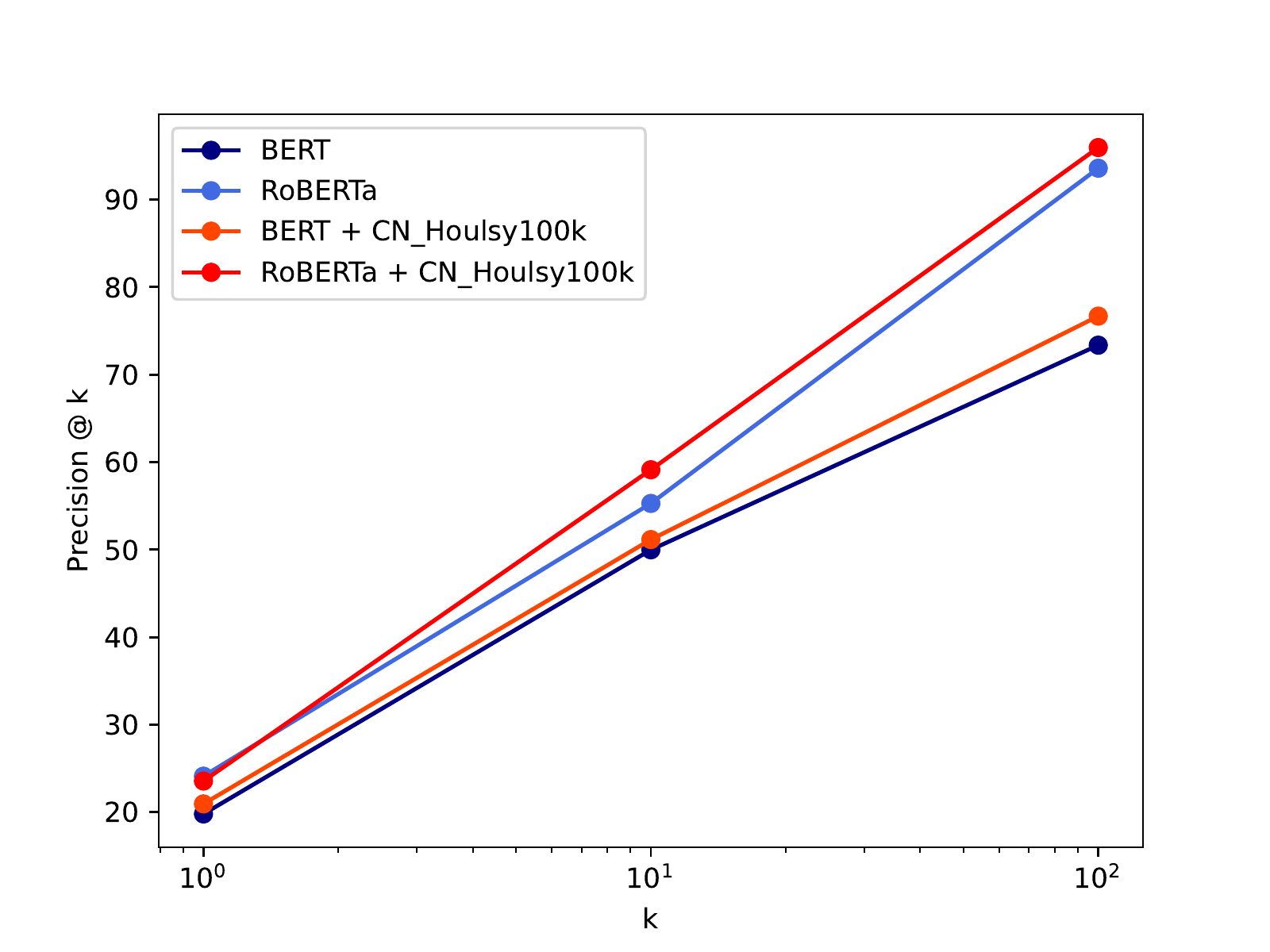}}}%
	\caption{Mean P@k curve for base models and the Houlsby adapter configuration. Base 10 log scale for the X axis. \textbf{a)} shows the result for all the predicates in the ConceptNet split of \textsc{lama} while \textbf{b) }shows results for the "\textsc{IsA}" predicate only}
	\label{fig:lamaAnalysis}%
\end{figure}

Figure \ref{fig:lamaAnalysis} shows the mean P@K curves for two language models, with and without an adapter. Part \textit{(a)} of the figure shows the result over all the predicate types present in the ConceptNet split of \textsc{lama} ($N=29774$). The injection of the adapter module decreases the performance of both \textsc{bert} and \textsc{RoBERTa} for all values of \textit{k}. However, the corpus with the Retrograph predicate set that adapters trained on only includes one of these types. Hence, there is little similarity between the two sets, and the reproduction of factual knowledge cannot be expected here. This also indicate that training on one set of predicate types does not improve the reproduction of facts on others.

Part \textit{(b)} of figure \ref{fig:lamaAnalysis}, on the other hand, shows the same models and adapters, but with the probe restricted only to the \textsc{IsA} predicate type --- which is then present both in the training corpus and in the probe. In the corpus from \citet{lauscher2020common}, triples with this predicate type make up 23\% of the total corpus ($N=69843$).

Since both resources are extracted from ConceptNet, we check the overlap between the masked tokens in the object position in \textsc{lama} and the object position in the triplets in the training set for the adapters. The actual percentage will depend on the random walk procedure, but for the sets used in figure \ref{fig:lamaAnalysis} there is a $5.7\%$ overlap between concepts. That is, approximately five percent of the concepts from \textsc{lama} that the models are expected to predict are also in the training corpus in some form, either with the same predicate type as in the probe, \textsc{IsA}, or one of the others in the Retrograph set.
 
Despite this, the injected models perform consistently better. As this performance gain is achieved by adding only 2.1\% additional parameters to the original model, and without adjusting the original weights at all, we interpret the results as an indication that this method of knowledge injection is effective. 

\section{Changing the predicate set}
\begin{figure}
	\begin{center}
		\includegraphics[width=0.5\textwidth]{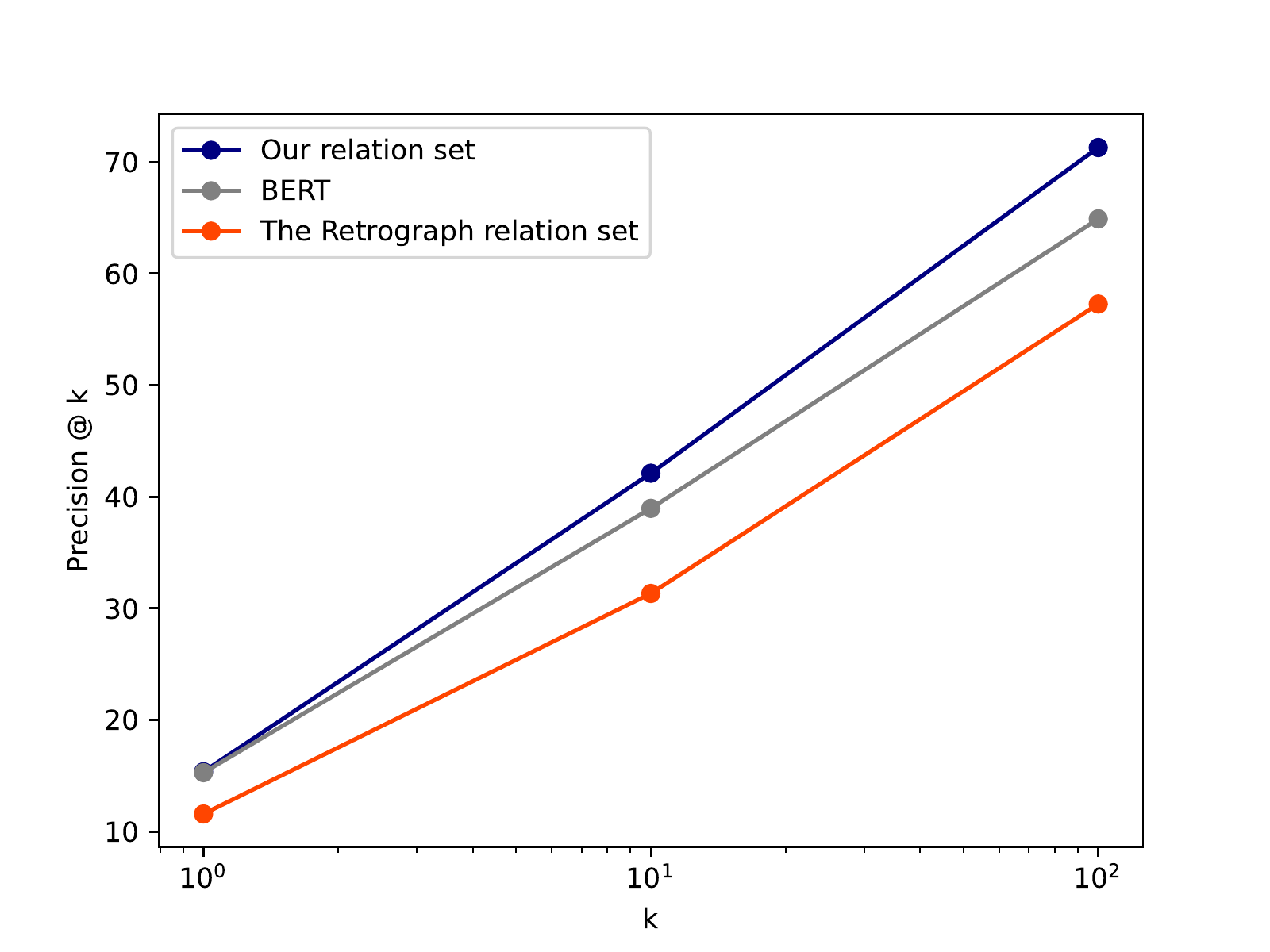}
	\end{center}
	\caption{The result of different training configurations on the ConceptNet split of LAMA \citep{petroni2019language}. The two models, in dark blue and orange, use \textsc{bert}\textsubscript{\textsc{base}} as the root model and the Houlsby configuration for their adapter, but are trained on different predicate sets of the ConceptNet graph. The gray line represents a \textsc{bert} model without any adapter training.}
	\label{fig:allLamaRelationsResult}
\end{figure}

In order to further probe the effectiveness of the proposed method, we introduce a new corpus ($N=99603$ triples) --- distilled with the same random walk procedure, but over a new set of predicate types, namely the same set of predicate types found in the ConceptNet split of \textsc{lama}. By intuition, if the method is effective, the injected models should score higher on average over all these predicate types than their non-injected counterparts. A list of these predicate types can be found in appendix \ref{sec:appendix}.
	
Figure \ref{fig:allLamaRelationsResult} compares the result of the injected models trained over our predicate set with that of the Retrograph set and a plain \textsc{bert} model for different values of k. As can be seen from the P@K curves, models trained over our predicate set improve the performance on the full ConceptNet split of the \textsc{LAMA} (N= 29774) probe by up to 6.39\% for \textsc{bert} at large values of k. For k=1, where the model must guess the correct masked object "at first try", we see little difference. Compared to the Retrograph set, which has fewer predicate types, the difference in performance indicate that predicate type specificity is important (e.g subgraph quality). For this comparison, the overlap between the training corpus for the adapters and the full ConceptNet split of \textsc{lama} is 36\% on the object level, meaning that roughly one third of the concepts were seen during training in some form.

This provides some evidence for the success of the knowledge injection. Models are able to reproduce factual knowledge when queried over the \textsc{lama} probe, even though the phrasing of the questions in \textsc{lama} is different than the strict triplet-style of the training corpus.

\section{Conclusion and Future Work}
Combining structured information and large pre-trained language models is a standing problem in NLP research. In this work, we show that training adapter modules on triplets extracted from ConceptNet using masked language modeling can help language models reproduce factual knowledge. Experiments on the ConceptNet split of the \textsc{lama} probe show that our adapter-injected models perform better in a zero-shot setting than non-injected models, having seen only a third of the relevant factual knowledge during pre-training in some form, encoded into only 2.1\% of the total parameters of the total model. Future work should investigate how this type of knowledge injection can augment language models on other types of tasks, such as language generation, multiple choice questions or natural language inference, which would require more fine-grained annotations of downstream tasks targeted at some form of knowledge. 

\bibliography{anthology,custom}

\begin{thebibliography}{23}
\expandafter\ifx\csname natexlab\endcsname\relax\def\natexlab#1{#1}\fi

\bibitem[{Devlin et~al.(2019)Devlin, Chang, Lee, and
  Toutanova}]{devlin-etal-2019-bert}
Jacob Devlin, Ming-Wei Chang, Kenton Lee, and Kristina Toutanova. 2019.
\newblock \href {https://doi.org/10.18653/v1/N19-1423} {{BERT}: Pre-training of
  deep bidirectional transformers for language understanding}.
\newblock In \emph{Proceedings of the 2019 Conference of the North {A}merican
  Chapter of the Association for Computational Linguistics: Human Language
  Technologies, Volume 1 (Long and Short Papers)}, pages 4171--4186,
  Minneapolis, Minnesota. Association for Computational Linguistics.

\bibitem[{Grover and Leskovec(2016)}]{grover2016node2vec}
Aditya Grover and Jure Leskovec. 2016.
\newblock \href {http://arxiv.org/abs/1607.00653} {node2vec: Scalable feature
  learning for networks}.

\bibitem[{Hendrycks and Gimpel(2020)}]{hendrycks2020gaussian}
Dan Hendrycks and Kevin Gimpel. 2020.
\newblock \href {http://arxiv.org/abs/1606.08415} {Gaussian error linear units
  (gelus)}.

\bibitem[{Houlsby et~al.(2019)Houlsby, Giurgiu, Jastrzebski, Morrone,
  De~Laroussilhe, Gesmundo, Attariyan, and
  Gelly}]{houlsby2019parameterefficient}
Neil Houlsby, Andrei Giurgiu, Stanislaw Jastrzebski, Bruna Morrone, Quentin
  De~Laroussilhe, Andrea Gesmundo, Mona Attariyan, and Sylvain Gelly. 2019.
\newblock \href {https://proceedings.mlr.press/v97/houlsby19a.html}
  {Parameter-efficient transfer learning for {NLP}}.
\newblock In \emph{Proceedings of the 36th International Conference on Machine
  Learning}, volume~97 of \emph{Proceedings of Machine Learning Research},
  pages 2790--2799. PMLR.

\bibitem[{Kaur et~al.(2022)Kaur, Bhatia, Aggarwal, Bansal, and
  Krishnamurthy}]{kaur-etal-2022-lm}
Jivat Kaur, Sumit Bhatia, Milan Aggarwal, Rachit Bansal, and Balaji
  Krishnamurthy. 2022.
\newblock \href {https://doi.org/10.18653/v1/2022.findings-naacl.57}
  {{LM}-{CORE}: Language models with contextually relevant external knowledge}.
\newblock In \emph{Findings of the Association for Computational Linguistics:
  NAACL 2022}, pages 750--769, Seattle, United States. Association for
  Computational Linguistics.

\bibitem[{Kingma and Ba(2017)}]{kingma2017adam}
Diederik~P. Kingma and Jimmy Ba. 2017.
\newblock \href {http://arxiv.org/abs/1412.6980} {Adam: A method for stochastic
  optimization}.

\bibitem[{Lauscher et~al.(2020)Lauscher, Majewska, Ribeiro, Gurevych, Rozanov,
  and Glava{\v{s}}}]{lauscher2020common}
Anne Lauscher, Olga Majewska, Leonardo F.~R. Ribeiro, Iryna Gurevych, Nikolai
  Rozanov, and Goran Glava{\v{s}}. 2020.
\newblock \href {https://doi.org/10.18653/v1/2020.deelio-1.5} {Common sense or
  world knowledge? investigating adapter-based knowledge injection into
  pretrained transformers}.
\newblock In \emph{Proceedings of Deep Learning Inside Out (DeeLIO): The First
  Workshop on Knowledge Extraction and Integration for Deep Learning
  Architectures}, pages 43--49, Online. Association for Computational
  Linguistics.

\bibitem[{Liu et~al.(2020)Liu, Zhou, Zhao, Wang, Ju, Deng, and
  Wang}]{Liu_Zhou_Zhao_Wang_Ju_Deng_Wang_2020}
Weijie Liu, Peng Zhou, Zhe Zhao, Zhiruo Wang, Qi~Ju, Haotang Deng, and Ping
  Wang. 2020.
\newblock \href {https://doi.org/10.1609/aaai.v34i03.5681} {K-bert: Enabling
  language representation with knowledge graph}.
\newblock \emph{Proceedings of the AAAI Conference on Artificial Intelligence},
  34(03):2901--2908.

\bibitem[{Liu et~al.(2019)Liu, Ott, Goyal, Du, Joshi, Chen, Levy, Lewis,
  Zettlemoyer, and Stoyanov}]{liu2019roberta}
Yinhan Liu, Myle Ott, Naman Goyal, Jingfei Du, Mandar Joshi, Danqi Chen, Omer
  Levy, Mike Lewis, Luke Zettlemoyer, and Veselin Stoyanov. 2019.
\newblock Roberta: A robustly optimized bert pretraining approach.
\newblock \emph{arXiv preprint arXiv:1907.11692}.

\bibitem[{Meng et~al.(2021)Meng, Liu, Clark, Shareghi, and
  Collier}]{meng-etal-2021-mixture}
Zaiqiao Meng, Fangyu Liu, Thomas Clark, Ehsan Shareghi, and Nigel Collier.
  2021.
\newblock \href {https://doi.org/10.18653/v1/2021.emnlp-main.383}
  {Mixture-of-partitions: Infusing large biomedical knowledge graphs into
  {BERT}}.
\newblock In \emph{Proceedings of the 2021 Conference on Empirical Methods in
  Natural Language Processing}, pages 4672--4681, Online and Punta Cana,
  Dominican Republic. Association for Computational Linguistics.

\bibitem[{Miller(1995)}]{miller1995wordnet}
George~A Miller. 1995.
\newblock Wordnet: a lexical database for {E}nglish.
\newblock \emph{Communications of the ACM}, 38(11):39--41.

\bibitem[{Peters et~al.(2019)Peters, Neumann, Logan, Schwartz, Joshi, Singh,
  and Smith}]{peters-etal-2019-knowledge}
Matthew~E. Peters, Mark Neumann, Robert Logan, Roy Schwartz, Vidur Joshi,
  Sameer Singh, and Noah~A. Smith. 2019.
\newblock \href {https://doi.org/10.18653/v1/D19-1005} {Knowledge enhanced
  contextual word representations}.
\newblock In \emph{Proceedings of the 2019 Conference on Empirical Methods in
  Natural Language Processing and the 9th International Joint Conference on
  Natural Language Processing (EMNLP-IJCNLP)}, pages 43--54, Hong Kong, China.
  Association for Computational Linguistics.

\bibitem[{Petroni et~al.(2019)Petroni, Rockt{\"a}schel, Riedel, Lewis, Bakhtin,
  Wu, and Miller}]{petroni2019language}
Fabio Petroni, Tim Rockt{\"a}schel, Sebastian Riedel, Patrick Lewis, Anton
  Bakhtin, Yuxiang Wu, and Alexander Miller. 2019.
\newblock \href {https://doi.org/10.18653/v1/D19-1250} {Language models as
  knowledge bases?}
\newblock In \emph{Proceedings of the 2019 Conference on Empirical Methods in
  Natural Language Processing and the 9th International Joint Conference on
  Natural Language Processing (EMNLP-IJCNLP)}, pages 2463--2473, Hong Kong,
  China. Association for Computational Linguistics.

\bibitem[{Pfeiffer et~al.(2020)Pfeiffer, R{\"u}ckl{\'e}, Poth, Kamath,
  Vuli{\'c}, Ruder, Cho, and Gurevych}]{pfeiffer-etal-2020-adapterhub}
Jonas Pfeiffer, Andreas R{\"u}ckl{\'e}, Clifton Poth, Aishwarya Kamath, Ivan
  Vuli{\'c}, Sebastian Ruder, Kyunghyun Cho, and Iryna Gurevych. 2020.
\newblock \href {https://doi.org/10.18653/v1/2020.emnlp-demos.7}
  {{A}dapter{H}ub: A framework for adapting transformers}.
\newblock In \emph{Proceedings of the 2020 Conference on Empirical Methods in
  Natural Language Processing: System Demonstrations}, pages 46--54, Online.
  Association for Computational Linguistics.

\bibitem[{Shi et~al.(2017)Shi, Li, Yang, Qi, Pan, and Zhou}]{shi2017semantic}
Longxiang Shi, Shijian Li, Xiaoran Yang, Jiaheng Qi, Gang Pan, and Binbin Zhou.
  2017.
\newblock Semantic health knowledge graph: semantic integration of
  heterogeneous medical knowledge and services.
\newblock \emph{BioMed research international}, 2017.

\bibitem[{Speer et~al.(2017)Speer, Chin, and Havasi}]{speer2017conceptnet}
Robyn Speer, Joshua Chin, and Catherine Havasi. 2017.
\newblock Conceptnet 5.5: An open multilingual graph of general knowledge.
\newblock In \emph{Thirty-first AAAI conference on artificial intelligence}.

\bibitem[{Suchanek et~al.(2007)Suchanek, Kasneci, and
  Weikum}]{suchanek2007yago}
Fabian~M Suchanek, Gjergji Kasneci, and Gerhard Weikum. 2007.
\newblock Yago: a core of semantic knowledge.
\newblock In \emph{Proceedings of the 16th international conference on World
  Wide Web}, pages 697--706.

\bibitem[{Sun et~al.(2021)Sun, Shi, Qi, and Zhang}]{sun2021jointlk}
Yueqing Sun, Qi~Shi, Le~Qi, and Yu~Zhang. 2021.
\newblock \href {http://arxiv.org/abs/2112.02732} {Jointlk: Joint reasoning
  with language models and knowledge graphs for commonsense question
  answering}.

\bibitem[{Wang et~al.(2018)Wang, Singh, Michael, Hill, Levy, and
  Bowman}]{wang2018glue}
Alex Wang, Amanpreet Singh, Julian Michael, Felix Hill, Omer Levy, and Samuel
  Bowman. 2018.
\newblock \href {https://doi.org/10.18653/v1/W18-5446} {{GLUE}: A multi-task
  benchmark and analysis platform for natural language understanding}.
\newblock In \emph{Proceedings of the 2018 {EMNLP} Workshop {B}lackbox{NLP}:
  Analyzing and Interpreting Neural Networks for {NLP}}, pages 353--355,
  Brussels, Belgium. Association for Computational Linguistics.

\bibitem[{Wang et~al.(2021)Wang, Tang, Duan, Wei, Huang, Ji, Cao, Jiang, and
  Zhou}]{wang2020kadapter}
Ruize Wang, Duyu Tang, Nan Duan, Zhongyu Wei, Xuanjing Huang, Jianshu Ji,
  Guihong Cao, Daxin Jiang, and Ming Zhou. 2021.
\newblock \href {https://doi.org/10.18653/v1/2021.findings-acl.121}
  {{K-Adapter}: {I}nfusing {K}nowledge into {P}re-{T}rained {M}odels with
  {A}dapters}.
\newblock In \emph{Findings of the Association for Computational Linguistics:
  ACL-IJCNLP 2021}, pages 1405--1418, Online. Association for Computational
  Linguistics.

\bibitem[{Yasunaga et~al.(2021)Yasunaga, Ren, Bosselut, Liang, and
  Leskovec}]{yasunaga-etal-2021-qa}
Michihiro Yasunaga, Hongyu Ren, Antoine Bosselut, Percy Liang, and Jure
  Leskovec. 2021.
\newblock \href {https://doi.org/10.18653/v1/2021.naacl-main.45} {{QA}-{GNN}:
  Reasoning with language models and knowledge graphs for question answering}.
\newblock In \emph{Proceedings of the 2021 Conference of the North American
  Chapter of the Association for Computational Linguistics: Human Language
  Technologies}, pages 535--546, Online. Association for Computational
  Linguistics.

\bibitem[{Zhang et~al.(2021)Zhang, Cai, Wang, Qiu, Yang, and
  He}]{zhang-etal-2021-smedbert}
Taolin Zhang, Zerui Cai, Chengyu Wang, Minghui Qiu, Bite Yang, and Xiaofeng He.
  2021.
\newblock \href {https://doi.org/10.18653/v1/2021.acl-long.457} {{SM}ed{BERT}:
  A knowledge-enhanced pre-trained language model with structured semantics for
  medical text mining}.
\newblock In \emph{Proceedings of the 59th Annual Meeting of the Association
  for Computational Linguistics and the 11th International Joint Conference on
  Natural Language Processing (Volume 1: Long Papers)}, pages 5882--5893,
  Online. Association for Computational Linguistics.

\bibitem[{Zhang et~al.(2022)Zhang, Bosselut, Yasunaga, Ren, Liang, Manning, and
  Leskovec}]{zhang2022greaselm}
Xikun Zhang, Antoine Bosselut, Michihiro Yasunaga, Hongyu Ren, Percy Liang,
  Christopher~D Manning, and Jure Leskovec. 2022.
\newblock \href {https://openreview.net/forum?id=41e9o6cQPj} {Grease{LM}: Graph
  {REAS}oning enhanced language models}.
\newblock In \emph{International Conference on Learning Representations}.

\end{thebibliography}
\bibliographystyle{acl_natbib}
\newpage
\appendix

\section{Appendix}
\label{sec:appendix}

The ConceptNet split of the \textsc{lama} probe includes the following predicate types:
\\
\\
\begin{example}
\textit{atLocation, capableOf, causes, causesDesire, desires, hasA, hasPrerequisite, hasProperty, hasSubevent, isA, locatedNear, madeOf, motivatedByGoal, partOf, receivesAction, usedFor.}
\end{example}

\subsection{Random walk procedure}
Retrograph uses the weighted random walk algorithm from \textsc{node2vec} \citep{grover2016node2vec} in order to extract the \texttt{subject---predicate---object} triples from ConceptNet. The pseudocode from the original publication on this algorithm is presented below. The alias method refers to a way of sampling from a discrete probability distribution.\footnote{\url{https://lips.cs.princeton.edu/the-alias-method-efficient-sampling-with-many-discrete-outcomes/}}

\begin{algorithm}
	\caption{The random walk procedure from Lauscher et al. (2020)}\label{node2vec}
	\begin{algorithmic}[1]
		\Procedure{node2vecWalk}{Graph G' = (V, E, $\pi$), Start node \textit{u}, Length \textit{l}}
		\State $\textit{Inititalize walk to [u]}$
		\For{\texttt{walk\_iter = 1 \textbf{to} \textit{l}}}
		\State $\textit{curr = walk[-1]}$
		\State $V_{curr} =  GetNeighbors(curr, G')$
		\State $s = AliasSample(V_{curr}, \pi)$
		\State $\textit{Append \textit{s} to \textit{walk}}$
		\EndFor
		\Return walk
		\EndProcedure
	\end{algorithmic}
\end{algorithm}

\end{document}